\documentclass[runningheads]{llncs}
\usepackage[T1]{fontenc}
\usepackage[hidelinks]{hyperref}
\usepackage{enumitem}
\usepackage{amsmath,amssymb}
\usepackage{array}
\usepackage{booktabs}
\usepackage{float}
\usepackage{multirow}
\usepackage{tabularx}
\usepackage{makecell}
\usepackage{caption}
\usepackage{chngcntr}
\usepackage{subcaption}

\usepackage{graphicx}
\usepackage{color}
\usepackage{acronym}
\acrodef{FEA}{Finite Element Analysis}
\acrodef{FE}{finite element}
\acrodef{PhysGNN}{Physics-based Graph Neural Networks}
\acrodef{BCT}{breast computed tomography}
\acrodef{GNN}{Graph Neural Networks}
\acrodef{DL}{deep learning}
\acrodef{JK}{Jumping Knowledge}
\acrodef{LODO}{Leave-one-deformation-out}
\acrodef{MRI}{magnetic resonance imaging}

\begin{document}
\title{Graph Neural Networks for modelling breast biomechanical compression}
\titlerunning{GNNs for modelling breast biomechanical compression}
% If the paper title is too long for the running head, you can set
% an abbreviated paper title here
%
\author{Hadeel Awwad\orcidID{0009-0007-4054-4546} \and
Eloy García\orcidID{0000-0002-0587-1919} \and
Robert Martí
\orcidID{0000-0002-8080-2710}}
%
%index{Awwad, Hadeel}
%index{García, Eloy}
%index{Martí, Robert}
% 
\authorrunning{H. Awwad et al.}
% First names are abbreviated in the running head.
% If there are more than two authors, 'et al.' is used.
%
\institute{Computer Vision and Robotics Institute, University of Girona, Girona, Spain\
\email{\{robert.marti\}@udg.edu}}
\maketitle              % typeset the header of the contribution
\begin{abstract}
Breast compression simulation is essential for accurate image registration from 3D modalities to X-ray procedures like mammography. It accounts for tissue shape and position changes due to compression, ensuring precise alignment and improved analysis. Although \ac{FEA} is reliable for approximating soft tissue deformation, it struggles with balancing accuracy and computational efficiency. Recent studies have used data-driven models trained on \ac{FEA} results to speed up tissue deformation predictions. We propose to explore \ac{PhysGNN} for breast compression simulation. \ac{PhysGNN} has been used for data-driven modelling in other domains, and this work presents the first investigation of their potential in predicting breast deformation during mammographic compression. Unlike conventional data-driven models, \ac{PhysGNN}, which incorporates mesh structural information and enables inductive learning on unstructured grids, is well-suited for capturing complex breast tissue geometries. Trained on deformations from incremental \ac{FEA} simulations, \ac{PhysGNN}'s performance is evaluated by comparing predicted nodal displacements with those from \ac{FE} simulations. This \ac{DL} framework shows promise for accurate, rapid breast deformation approximations, offering enhanced computational efficiency for real-world scenarios.

\keywords{Breast computed tomography  \and Digital breast phantom \and Multimodal imaging fusion \and 2D-3D image registration \and Mammography \and Finite element models}
\end{abstract}
\section{Introduction}

Breast cancer is one of the most prevalent and, life-threatening cancers affecting women worldwide. The high incidence rate underscores the critical need for early and accurate diagnosis to improve patient outcomes \cite{sung2021,zhang2023}. Common diagnostic modalities include mammography, \ac{MRI}, ultrasound, and dedicated \ac{BCT}, each offering unique benefits in detecting and characterizing breast lesions \cite{Caballo2021,Tan2023,Zhang2021PredictingMS,Zhang2023a}. Integrating and correlating information from these diverse imaging techniques can significantly enhance diagnostic accuracy. By synthesizing and simulating data from 3D modalities like \ac{MRI} and \ac{BCT} into 2D representations as seen in mammography, clinicians can leverage the comprehensive structural details of \ac{MRI} or \ac{BCT} with the practical format of mammography. This multi-modal strategy provides a more comprehensive perspective of breast tissue, enhancing lesion detection and assessment, and ultimately improving diagnostic outcomes and patient care.

Achieving realistic breast compression simulation is crucial for multi-modal breast imaging correspondence, commonly approached using Finite Element Analysis (\ac{FEA}) methods. However, the limited computational efficiency of \ac{FEA} poses a significant challenge in medical image registration, as it requires substantial time and resources to solve biomechanical models accurately. This limitation affects the performance of existing multimodal registration techniques. To address this issue, researchers have proposed data-driven models that train different machine learning algorithms with \ac{FEA} results to speed up tissue deformation approximations by prediction \cite{MARTINEZ2017,mendizabal2020physics,phellan2021real,ruperez2018modeling,said2023}. However, these methods overlook valuable information contained within the \ac{FE} mesh structure, such as node connections and distances, and their number of parameters depends on the mesh resolution.

To the best of our knowledge, this study is the first to adapt the \ac{PhysGNN} model proposed by Salehi \cite{physgnn2022} for simulating breast compression. It provides a comparative analysis of the results using both quantitative and qualitative metrics against existing \ac{FEA} methods and discusses its benefits and limitations.

\section{Materials}
\begin{figure}[h!]
    \centering
    \begin{subfigure}[b]{0.4\textwidth}
        \centering
        \includegraphics[width=\textwidth]{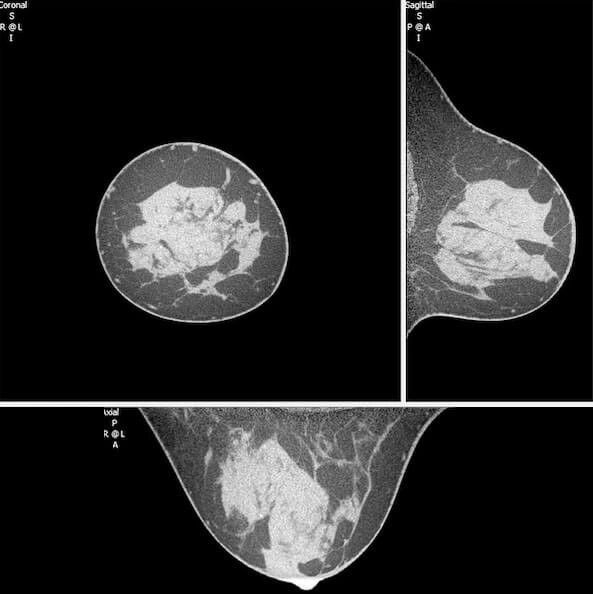} 
        \caption{}
        \label{fig:bct}
    \end{subfigure}
    \hfill
    \begin{subfigure}[b]{0.4\textwidth}
        \centering
        \includegraphics[width=\textwidth]{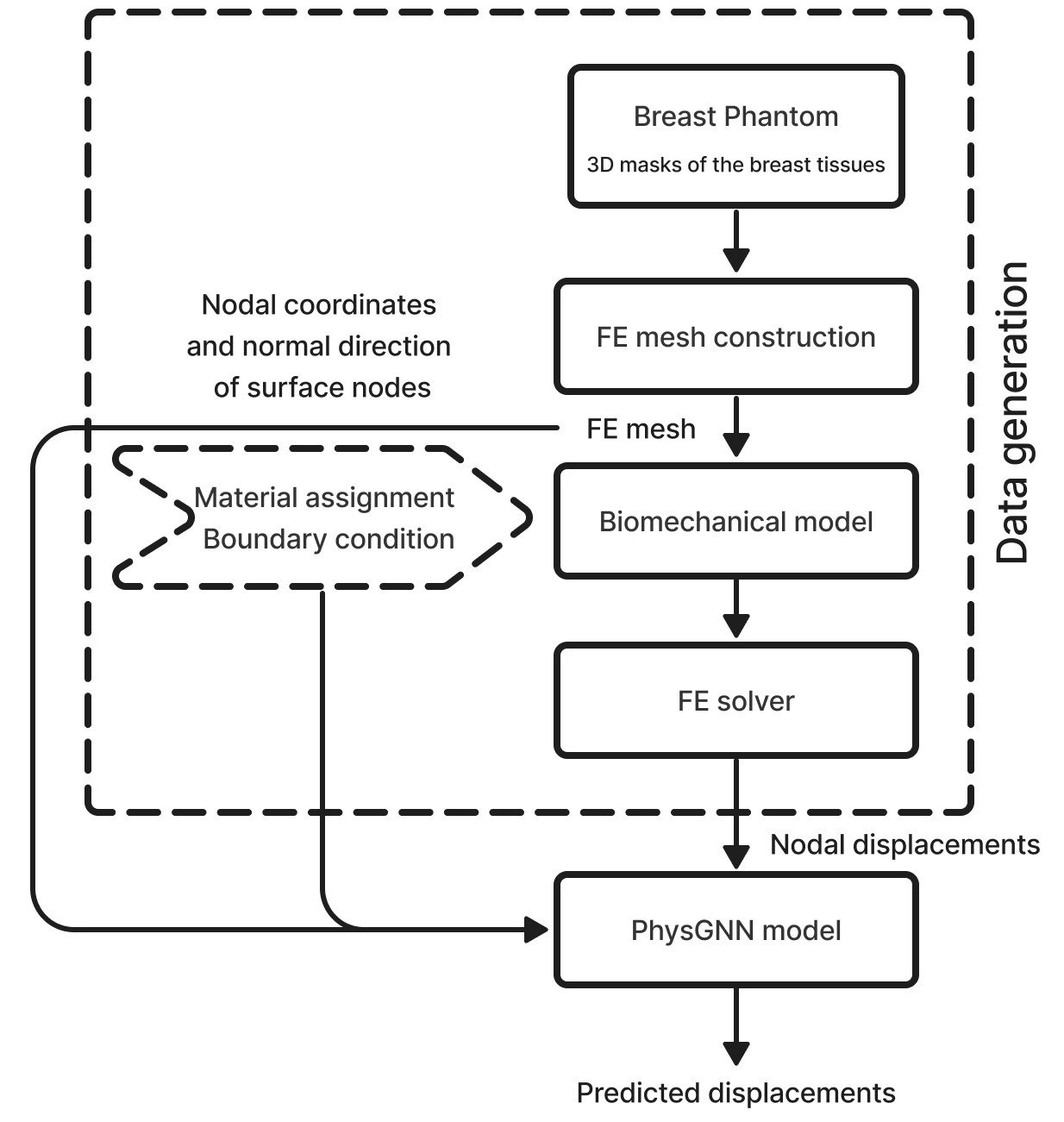} 
        \caption{}
        \label{fig:method}
    \end{subfigure}
    \caption{(a) Reconstructed dataset from breast CT \cite{gazi2015evolution}. (b) Flow chart of the proposed method to simulate the breast compression using PhysGNN.}
    \label{fig:combined}
\end{figure}

\ac{BCT} is an advanced imaging modality that provides high-resolution, three-dimensional breast images, offering superior tissue contrast compared to digital mammography, see Figure~\ref{fig:combined} (a). This enhanced imaging capability facilitates better detection and characterization of breast lesions. This work used a publicly available dataset of computational digital phantoms generated from clinical \ac{BCT} images previously obtained at UC Davis (California, USA) \cite{bct2001,gazi2015evolution,tomo2021}, utilizing a semi-automatic tissue classification algorithm \cite{seg2020}. 
Each voxel of the breast phantom was segmented into various classes: air, fatty tissue, glandular tissue, and skin tissue. In this work, we have used a single phantom (UncompressedBreast3) for generating the geometry of the biomechanical model. We chose to use a single phantom because generating incremental \ac{FEA} solutions is computationally expensive. Future research can explore additional breast phantoms to test \ac{PhysGNN}'s capability in predicting the deformation of various geometries.

\section{Methods}
Figure~\ref{fig:combined} (b) outlines the workflow of the proposed method to approximate mammographic compression using a biomechanical breast model to train the PhysGNN model, detailed in the following sections.

\subsection{Biomechanical Breast Model}
\begin{figure}[!ht]
    \centering
    \includegraphics[width=1.0\columnwidth]{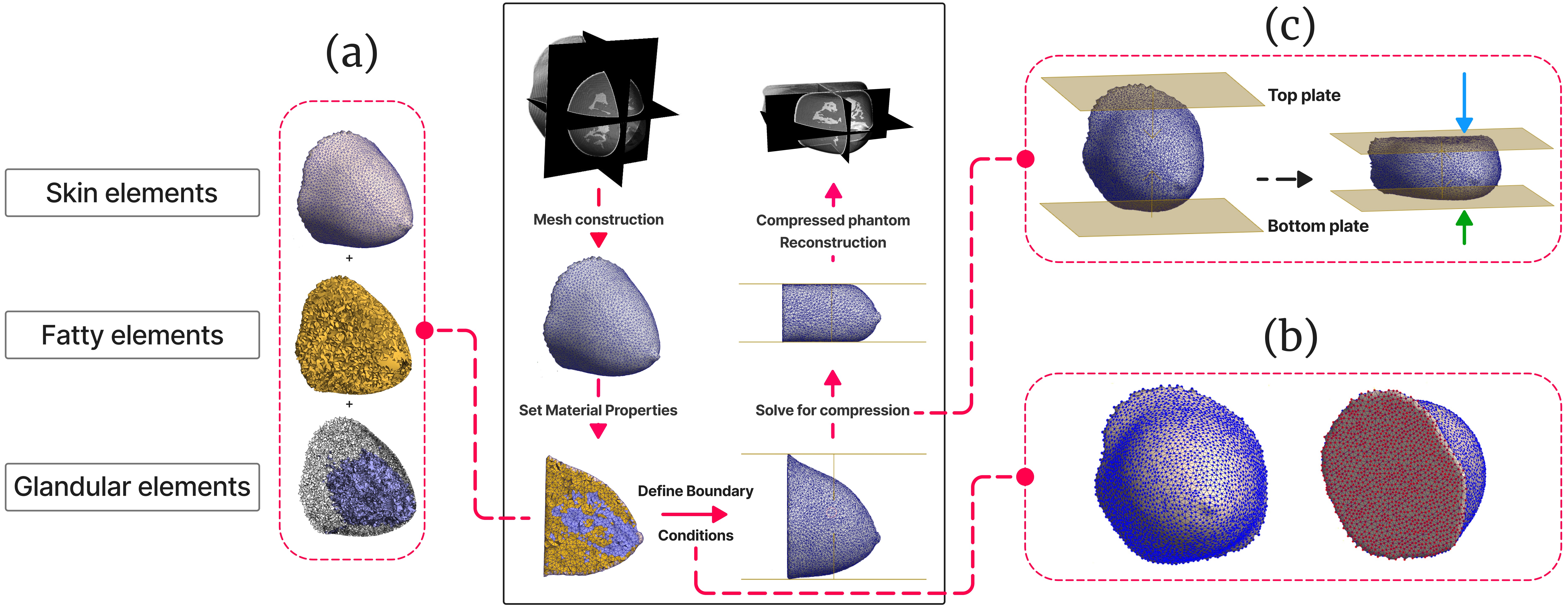} 
    \caption{Overview of steps to set up and solve compressed breast geometry using FEA, then apply deformation to reposition and compress the breast phantom.}
    \label{fig:overview}
\end{figure}

The breast biomechanical model is derived from García's work \cite{bct1}. An overview of generating a biomechanical breast model is shown in Figure~\ref{fig:overview} involving breast geometry and mesh generation, material properties and boundary conditions definition, and the finite element solver.

\textbf{Generation of the Finite Element Mesh} To represent the breast geometry, we used a meshing technique that discretizes the breast volume into small tetrahedral elements \cite{Jonathan1970}. These elements form a graph-like structure, approximating the breast's complex geometry with tetrahedrons, each having four triangular faces and four nodes. The \ac{FE} volume mesh for running \ac{FE} simulations was generated by running Pygalmesh \cite{pygalmesh}, a Python interface to the CGAL library \cite{cgal} to generate high-quality 3D volume mesh. Figure~\ref{fig:overview} (a) illustrates the \ac{FE} volume mesh of the breast phantom, which consists of 17595 nodes and 95865 tetrahedral elements. More details on the \ac{FE} volume mesh may be found in Table~\ref{tab:statmesh}, Appendix 1.

\textbf{Material Properties and Boundary Conditions}
The material behaviour of breast tissues was characterized as nearly incompressible, with a Poisson ratio of 0.49, using homogeneous and isotropic Neo-Hookean material models as reported in the literature \cite{neohok}. The stiffness measures were assigned using Young's modulus (\( E_{\text{fatty}} = 4.46 \, \text{kPa} \), \( E_{\text{glandular}} = 15.1 \, \text{kPa} \), \( E_{\text{skin}} = 20.0 \, \text{kPa} \)). 
The boundary condition was set to restrict rigid motion by constraining the posterior surface of the breast in the anterior-posterior direction. See Figure~\ref{fig:overview} (b), which is colour-coded by boundary conditions: free nodes (blue) throughout the breast and constrained nodes (red) on the posterior surface to mimic chest wall restriction.

\textbf{Finite Element Simulations}
While \ac{FEA} is a common method for solving the compression, this work specifically focuses on simulating compression for the Cranio Caudal (CC) view during mammography. The results from \ac{FEA} simulations are utilized as ground truth to train the \ac{PhysGNN} model in predicting breast compression, see next section. NiftySim (v.2.3.1; University College London, UK) \cite{Johnsen2015} is used to conduct incremental \ac{FE} simulations on the \ac{FE} mesh, modelled using two frictionless infinite linear planes \cite{bct1}, see Figure~\ref{fig:overview} (c).

\subsection{PhysGNN}

\begin{figure*}[t]
  \centering
  \includegraphics[width=1.0\columnwidth]{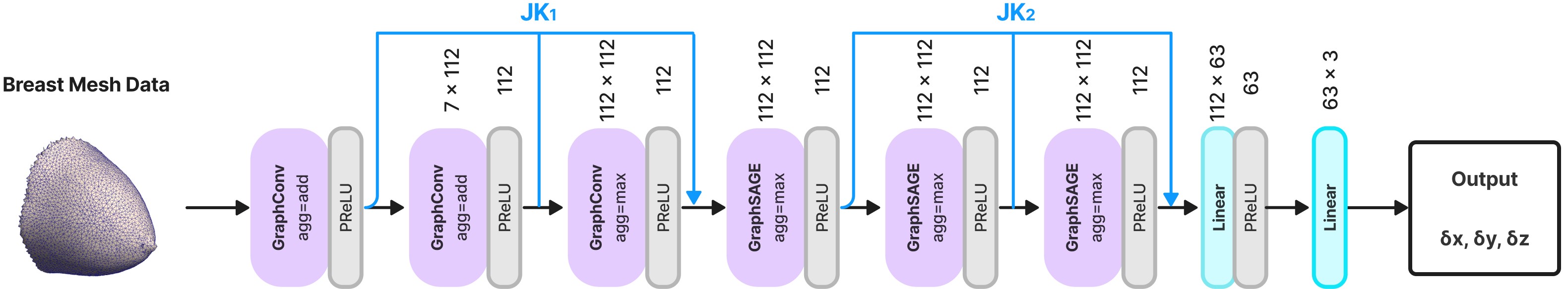}  
  \caption{Architectural diagram of \ac{PhysGNN} \cite{physgnn2022}}
  \label{fig:arch}
\end{figure*}

In this work, we explore \ac{PhysGNN} proposed by Salehi \cite{physgnn2022}, which approximates soft tissue deformation under prescribed forces by leveraging \ac{GNN}s. Figure~\ref{fig:arch} illustrates the architecture of \ac{PhysGNN}, integrating GraphSAGE and GraphConv layers with \ac{JK} connections. This design facilitates inductive learning over graphs through \textit{parameter sharing}, ensuring a constant number of parameters regardless of the mesh size. This approach enhances computational feasibility when learning from high-quality meshes with numerous nodes \cite{physgnn2022}. 

A training and testing dataset was generated from the incremental simulations of NiftySim by applying a force of 90 Newtons to the breast surface nodes, incrementally over 30-time steps and in 40 directions, to capture the non-linear behaviour of soft tissue under large forces. The forces applied to each surface node are directed along its surface normal ($x$, $y$, and $z$) and three additional randomly sampled directions. Each additional direction is represented as a tuple ($x$, $y$, and $z$) randomly selected from a unit-radius hemisphere. Note that this is slightly different from the original implementation of \cite{physgnn2022} where only 10 distinct batches of random directions were included, along with the surface normal direction. Further details on the generated dataset can be found in Table~\ref{tab:dataset}, Appendix 2.

The input features for \ac{PhysGNN} include force values applied to surface nodes in Cartesian coordinates $(F_x, F_y, F_z)$ and spherical coordinates $(F_\rho, F_\theta, F_\phi)$. Each node is also assigned a Physical Property value: 0.1 for skin, 0.6 for glandular tissue, 1 for fatty tissue under free boundary conditions, and 0 for fixed boundary conditions. This value impacts the node's displacement, influenced by its boundary condition and Young's modulus. Fatty tissue tends to experience larger displacements due to its smaller Young’s modulus compared to glandular and skin tissues. \ac{PhysGNN} outputs displacement values $(\delta_x, \delta_y, \delta_z)$ for the mesh nodes. Edge weights $(e_{u,v})$ in the \ac{GNN} model are computed as the inverse of the Euclidean distance between adjacent nodes $u$ and $v$, with $u, v \in  \mathcal{N}$. Information about \ac{PhysGNN} hyperparameters and infrastructure settings can be found in Appendix 3.

\subsection{Voxelised Compressed Phantom Reconstruction}
Predicted deformations by \ac{FEA} and \ac{PhysGNN} are used to reconstruct the compressed voxel phantom following the method in \cite{mapping2018,mapping2017}. The compressed FE mesh is converted into a voxelized phantom, with elements stored on a uniform grid. Each voxel's position is mapped to the uncompressed model using barycentric coordinates \cite{mapping2018}. This transforms points along a ray in the compressed model to a curve in the segmented \ac{BCT} image, with labels obtained using nearest neighbour interpolation.

\subsection{PhysGNN Training Experiments}
We evaluate the performance of \ac{PhysGNN} model on Hold-out and \ac{LODO} strategies against the results of \ac{FEA} (our baseline). 

\textbf{Hold-out experiment}
The dataset from step-wise compression of a single \ac{BCT} phantom obtained using \ac{FEA} NiftySim was randomly split into training (70\%), validation (20\%), and testing (10\%) sets. Despite using a single breast phantom with 30 deformation states, this experiment provides valuable groundwork for future studies involving additional breast models.

\textbf{Leave-one-deformation-out experiment} To evaluate the model's ability to generalize to unseen deformations, a specific testing strategy was used. From 30 NiftySim-simulated deformations, one (the final compression state) was isolated for testing. The remaining 29 were split into training (80\%) and validation (20\%) sets. This approach reflects real-world scenarios where models train on known deformations and predict new ones for a complete breast geometry. Note that the training data is from a single \ac{BCT} phantom.

\subsection{Evaluation Metrics} 

Similar to previous research \cite{MARTINEZ2017,said2023}, the \ac{PhysGNN} model predictions are evaluated against \ac{FEA} results, which serve as the ground truth. We use several evaluation metrics to assess model performance and accuracy. The Dice coefficient measures the agreement between segmented images. The Mean Euclidean Error (MEE) and Mean Absolute Error (MAE) assess prediction accuracy based on node positions, with MEE measuring the average distance between predicted and ground truth positions, and MAE evaluating the average error magnitude of node displacements. Lastly, the Volume Loss metric measures the difference in breast tissue volume before and after compression, reflecting the physical realism and accuracy of the biomechanical models. These metrics collectively provide a comprehensive assessment of the model's performance.

\section{Results and Discussion}

\subsection{Performance of PhysGNN Model}

Table~\ref{tab:performance} shows PhysGNN's performance in predicting breast deformation under mammographic compression. \ac{PhysGNN} effectively predicts deformations, with 99.96\% of nodal position errors under 1 mm in the Hold-out experiment and 81.22\% in the \ac{LODO} experiment. The Hold-out evaluation offers more stable performance estimates than \ac{LODO}, which is computationally expensive and limits model complexity and hyperparameter tuning. Additionally, single-deformation testing in \ac{LODO} may not accurately reflect performance on diverse deformations. Table~\ref{tab:statistics} in Appendix 4 presents the \ac{PhysGNN} test set statistics.

\renewcommand{\arraystretch}{2.4}
\newcolumntype{C}[1]{>{\centering\arraybackslash}p{#1}}

\begin{table*}[!ht]
    \centering
    \caption{The performance of \ac{PhysGNN} model on Hold-out and Leave-one-deformation-out on the test set. Exp. stands for Experiment}
    \small  % Make the font of the table smaller
    % \vspace{6pt}
    \begin{tabularx}{\textwidth}{C{1.2cm}|C{1.4cm}|C{1.4cm}|C{1.4cm}|C{1.4cm}|C{1.4cm}|C{1.5cm}|C{1.6cm}}
        \hline
        \multirow{2}{*}{Exp.} & \multirow{2}{*}{\shortstack{MAE\\($\delta x$)\\(mm)}} & \multirow{2}{*}{\shortstack{MAE\\($\delta y$)\\(mm)}} & \multirow{2}{*}{\shortstack{MAE\\($\delta z$)\\(mm)}} & \multirow{2}{*}{\shortstack{Mean\\Euclidean\\Error\\(mm)}} & \multirow{2}{*}{\shortstack{Euclidean\\Error $\le$ \\1 mm\\(\%)}} & \multirow{2}{*}{\shortstack{Mean\\Absolute\\Position\\Error\\(mm)}} & \multirow{2}{*}{\shortstack{Absolute\\Position\\Error $\le$ \\1 mm\\(\%)}} \\
        & & & & & & & \\
        \hline        
        Hold-out & 0.17 $\pm$0.18 & 0.20 $\pm$0.20 & 0.13 $\pm$0.14 & 0.34 $\pm$0.15 & 97.50 & 0.17 $\pm$0.03 & 99.96 \\
        
        \ac{LODO} & 0.56 $\pm$0.44 & 0.52 $\pm$0.39 & 0.67 $\pm$0.52 & 1.21 $\pm$0.44 & 33.51 & 0.59 $\pm$0.07 & 81.22 \\
        \hline
    \end{tabularx}
    \label{tab:performance}
\end{table*}

In the Hold-out experiment, predicting tissue deformation took 0.42 ± 0.04 seconds on CPU and 0.01 ± 0.06 seconds on GPU. In the \ac{LODO} experiment, it took 0.82 seconds on CPU and 0.47 seconds on GPU. Incremental NiftySim simulations, with GPU acceleration, totalled 4640.5 seconds (154.7 seconds per simulation). This indicates a speedup of 329 times with GPU and 188 times with CPU in the \ac{LODO} experiment compared to a single NiftySim simulation.

\subsection{Quantitative and Qualitative Analysis of Leave-one-deformation-out experiment}
We present here additional quantitative and qualitative results only for the \ac{LODO} experiment. For the Hold-out experiment, since graphs from different deformations are mixed in the test set, a specific evaluation cannot be provided.

\begin{figure*}[h!]
  \centering
  \includegraphics[width=0.75\columnwidth]{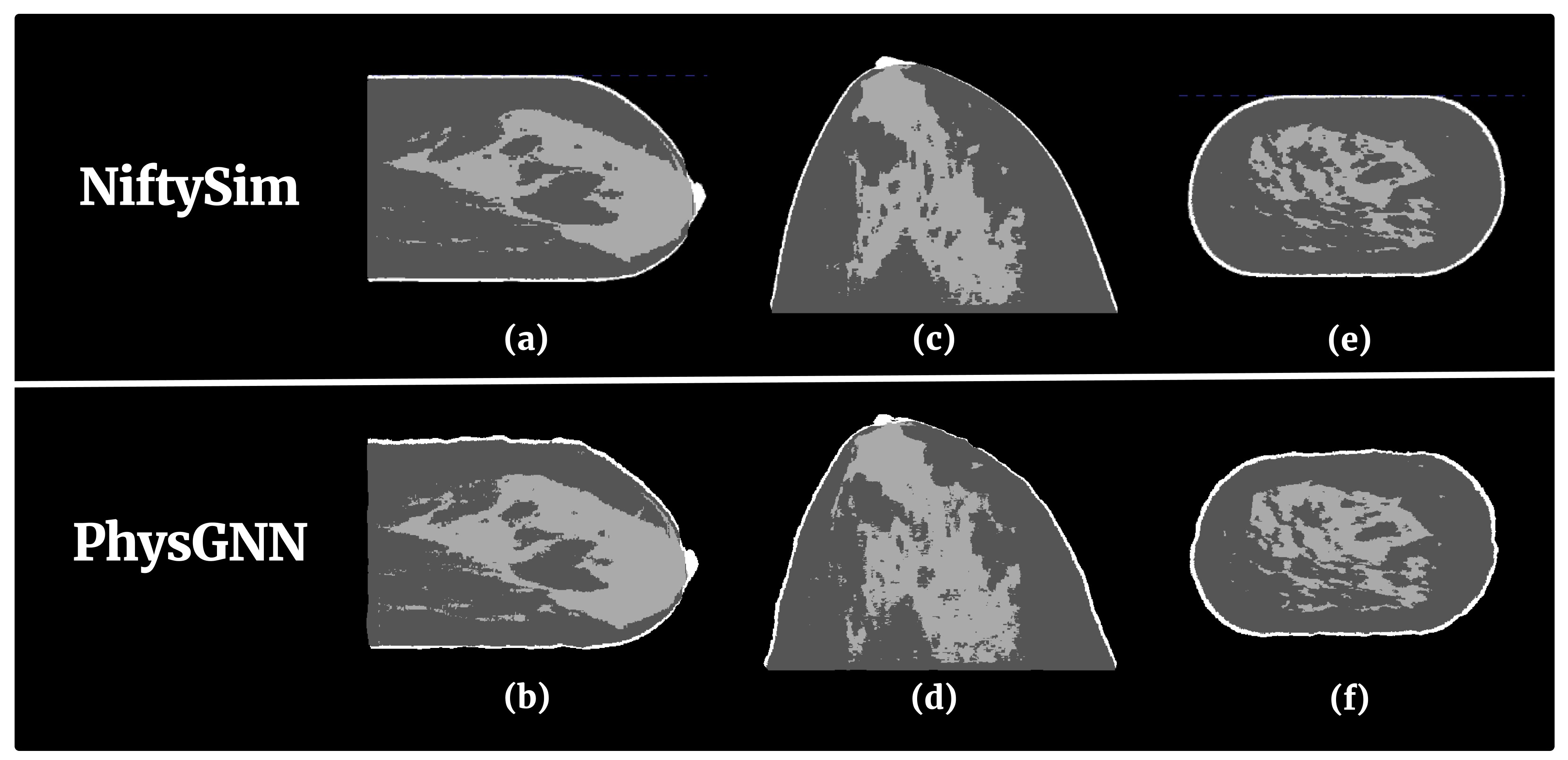}  
  \caption{Cross-sectional views of compressed digital phantoms from NiftySim and \ac{PhysGNN}: (a) and (b) sagittal sections, (c) and (d) axial sections, (e) and (f) coronal sections}
  \label{fig:cross-ph}
\end{figure*}

Figure~\ref{fig:cross-ph} compares the reconstructed phantoms from PhysGNN-predicted deformation using the \ac{LODO} strategy to the final compression state by NiftySim, which serves as the ground truth. The results show that PhysGNN closely matches \ac{FEA} in breast deformation predictions, demonstrating its effectiveness and potential as a reliable alternative to traditional \ac{FEA} methods. The similarity is further quantified by the Dice scores: 0.94 for fat tissue, 0.83 for glandular tissue, and 0.53 for skin tissue, indicating high values for primary tissues (excluding skin, which is very thin and thus results in a compromised Dice score).

% \begin{table}[h!]
%   \centering
%   \caption{Dice score of reconstructed compressed phantoms obtained from NiftySim and PhysGNN}
%   \small  % Make the font of the table smaller
%   % \vspace{6pt}
%   \begin{tabularx}{\linewidth}{>{\centering\arraybackslash}X >{\centering\arraybackslash}X >{\centering\arraybackslash}X}
%     \hline
%     \makecell{Fat Tissue} & \makecell{Glandular Tissue } & \makecell{Skin Tissue } \\
%     \hline
%     0.94 & 0.83 & 0.53 \\
%     \hline
%   \end{tabularx}
%   \label{tab:dice}
% \end{table}

Finally, Table~\ref{tab:loss} summarizes the volume loss percentages for total breast, fatty, glandular, and skin tissues. \ac{PhysGNN} predictions show a slightly increased overall breast tissue loss.

\newcolumntype{Y}{>{\centering\arraybackslash}X}
\renewcommand{\arraystretch}{1.0}

\begin{table*}[h!]
    \centering
    \caption{Breast tissue loss during compression}
    \small  % Make the font of the table smaller
    % \vspace{6pt}
    \begin{tabularx}{\textwidth}{c|YYYYY}
        \hline
        \multirow{2}{*}{\ac{FEA}/\ac{DL}} & \multirow{2}{*}{\shortstack{Total breast \\ volume loss(\%)}} & \multirow{2}{*}{\shortstack{Fatty tissue \\ volume loss(\%)}} & \multirow{2}{*}{\shortstack{Glandular tissue \\ volume loss(\%)}} & \multirow{2}{*}{\shortstack{Skin tissue \\ volume loss(\%)}} \\
        & & & &  \\
        \hline        
        NiftySim & 1.03 & 0.38 & 0.03 & 10.33 \\
        
        \shortstack{PhysGNN} & 1.26 & 0.55 & 0.34 & 10.94 \\
        \hline
    \end{tabularx}
    \label{tab:loss}
\end{table*}

\begin{figure*}[t]
  \centering
  \includegraphics[width=1.0\columnwidth]{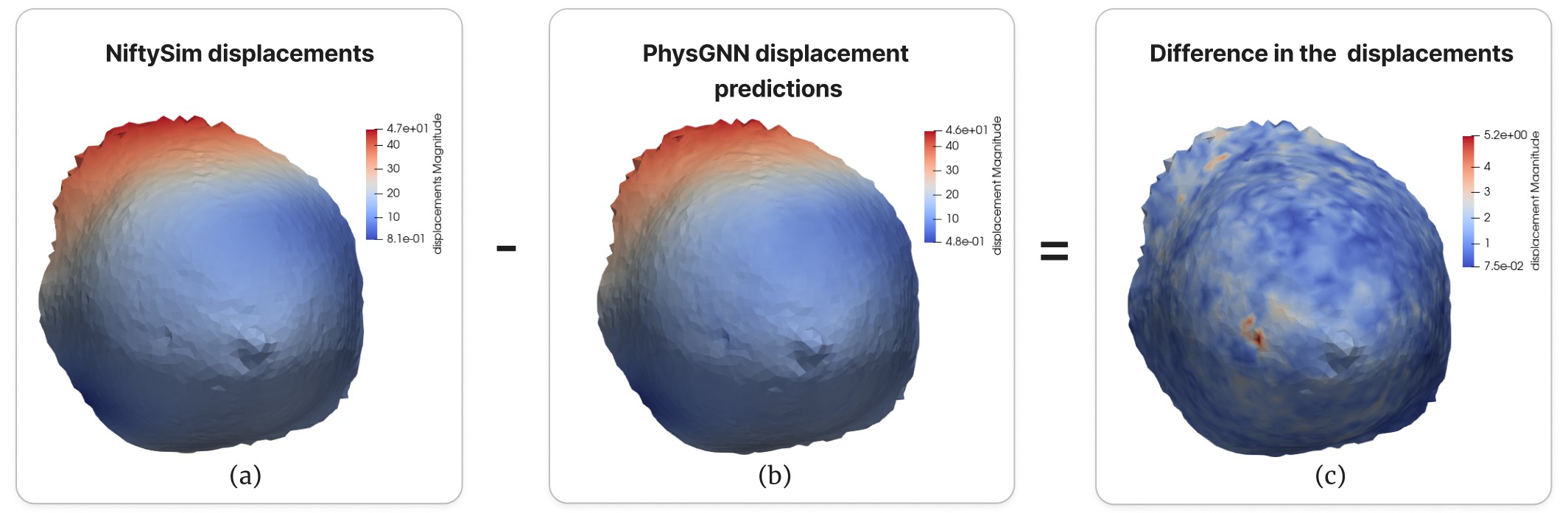}  
  \caption{Output displacements (\text{mm}) of NiftySim and PhysGNN on the FE model}
  \label{fig:displacement}
\end{figure*}

We can also visually assess the displacement magnitudes on the \ac{BCT} model with each method as shown in Figure~\ref{fig:displacement}. Figure~\ref{fig:displacement} (a) shows NiftySim's displacements, Figure~\ref{fig:displacement} (b) shows PhysGNN's predicted displacements, and Figure~\ref{fig:displacement} (c) highlights the differences. While overall patterns are similar, noticeable dissimilarity exists on the breast surface, with a maximum displacement difference of 5.2 mm.

The overall results show strong performance with low errors and visual similarity to the NiftySim ground truth, suggesting the model is nearing practical applicability. However, the necessity for training on incremental \ac{FEA} simulations, which requires significant training time, remains a challenge. Despite this, the method holds promise for multiple compression simulations, particularly within an optimization framework for 2D/3D registration, as demonstrated by García's work \cite{GARCIA201976}. Thus, PhysGNN has significant potential for specialized scenarios requiring detailed and repeated simulations.

\section{Conclusion}
\label{sec:Conclusion}

This work is the first to apply \ac{GNN}s for predicting breast deformation during mammographic compression, comparing quantitatively and qualitatively with standard \ac{FEA} models. The \ac{PhysGNN} model, trained on \ac{FEA} displacement data, achieved a Mean Euclidean Error of 0.34 ± 0.15 mm in the Hold-out experiment. The \ac{PhysGNN} model achieved accurate results comparable to the \ac{FEA}, with a speedup of factor 329 and 188 using GPU and CPU, respectively in the Leave-one-deformation-out experiment, suggesting its potential to replace \ac{FEA} simulations for real-time clinical applications. For future work, it is important to include simulations of compression for the Mediolateral Oblique (MLO) view during mammography. Further efforts should focus on comparing \ac{PhysGNN} with other \ac{FEA} methods to assess relative performance. Additionally, the study should be extended to generalize the model to multiple geometries, examining its robustness and accuracy across diverse scenarios. Finally, integrating \ac{PhysGNN} into 2D/3D registration frameworks could further enhance its applicability and effectiveness in clinical settings.
\\
\textbf{Acknowledgments.} 
This work has been partially funded by the Erasmus+: Erasmus Mundus Joint Master’s Degree (EMJMD) scholarship (2022–2024), with project reference 610600-EPP-1-2019-1-ES-EPPKA1-JMD-MOB and the project VICTORIA, “PID2021-123390OB-C21” from the Ministerio de Ciencia e Innovación of Spain.
\\
\textbf{Disclosure of Interests.}
The authors have no competing interests to declare relevant to this article's content.

%
% ---- Bibliography ----
%
% BibTeX users should specify bibliography style 'splncs04'.
% References will then be sorted and formatted in the correct style.
%
% \renewcommand{\bibname}{References}
\bibliographystyle{splncs04}
\bibliography{Paper-0042}
%
% \newpage
\appendix

% Redefine \thesection to use numbers instead of letters
\renewcommand\thesection{\arabic{section}}

\section*{Appendix 1}
\label{appendix:appendixb}
Table~\ref{tab:statmesh} provides information on the finite element (FE) volume mesh that was used for generating
the dataset.

\begin{table}[ht]
\centering
\caption{Finite element volume mesh statistics}
\begin{tabular}{c|c}
\hline
\textbf{Attribute} & \textbf{Value} \\
\hline
Mesh points & 17595 \\
\hline
Mesh tetrahedra & 95865 \\
\hline
Mesh triangular faces & 33594 \\
\hline
Mesh faces on the exterior surface & 2300 \\
\hline
\end{tabular}
\label{tab:statmesh}
\end{table}

\section*{Appendix 2}
\label{appendix:appendixe}

Further details on the generated training and testing datasets can be found in Table~\ref{tab:dataset}.

\begin{table}[ht]
\centering
\caption{Characteristics of the Dataset. B.C. stands for Boundary Condition}
\begin{tabular}{cccccc}
\hline
\multirow{4}{*}{\shortstack{No.\\Fixed\\B.C.\\Nodes}} & \multirow{4}{*}{\shortstack{No. Free\\B.C.\\Surface\\Nodes}} & \multirow{4}{*}{\shortstack{No. Force\\magnitudes}} & \multirow{4}{*}{\shortstack{Max.\\Force\\Applied\\(N)}} & \multirow{4}{*}{\shortstack{No.\\Directions}} & \multirow{4}{*}{\shortstack{No.\\Simulations}} \\
 &  &  &  &  \\
 &  &  &  &  \\
 &  &  &  &  \\
\hline
1171 & 1129 & 30 & 90 & 40 & 1200 \\
\hline
\end{tabular}
\label{tab:dataset}
\end{table}

\section*{Appendix 3}
\label{appendix:appendixc}

\textbf{PhysGNN Hyperparameters}
The loss function used for learning the trainable parameters is the mean Euclidean error computed as:

\begin{equation}
MEE = \frac{1}{\mathcal{N}} \sum_{n \in \mathcal{N}} \sqrt{\sum_{i=1}^{3} \left( y_{n}^{i} - \hat{y}_{n}^{i} \right)^2}
\end{equation}

where $\mathcal{N}$ is the number of mesh nodes, $\mathbf{y} \in \mathbb{R}^{\mathcal{N} \times 3}$ represents the FEM-approximated displacement in the $x$, $y$, and $z$ directions, and $\mathbf{\hat{y}} \in \mathbb{R}^{\mathcal{N} \times 3}$ represents the displacement predicted by \ac{PhysGNN}. The AdamW optimizer, with an initial learning rate of 0.005, was used to minimize the loss value, reducing the rate by a factor of 0.1 to a minimum of $1 \times 10^{-8}$ if validation loss did not improve after 5 epochs. Early stopping, halting training after 15 epochs without validation loss improvement, was employed to prevent overfitting. Additionally, a dropout rate of 0.1 was applied to the penultimate layer of \ac{PhysGNN} to enhance generalization. The model was trained in 8 batches for faster convergence.
\\
\textbf{Infrastructure Settings} 
The FE simulations and PhysGNN model training in our study were carried out on an NVIDIA GeForce RTX 2080 Ti GPU with 46 GB.

\section*{Appendix 4}
\label{appendix:appendixd}
Table~\ref{tab:statistics} presents the PhysGNN test set statistics of Hold-out and Leave-one-deformation-out experiments.

\renewcommand{\arraystretch}{1.8}

\begin{table*}[h!]
    \centering
    \caption{The test set statistics of Hold-out and Leave-one-deformation-out,  where \( y \) is the displacement, and \(\text{Max. Euclidean Error}_{\text{mean}}\) is computed as the average of the maximum Euclidean error observed for each data element—i.e., each simulation.}
    \small  % Make the font of the table smaller
    % \vspace{6pt}
    \begin{tabularx}{\textwidth}{c|YYY}
        \hline
        \multirow{2}{*}{Experiment} & \multirow{2}{*}{\shortstack{\(\delta y_{\text{max}} \, \text{(mm)}\)}} & \multirow{2}{*}{\shortstack{\(\delta y_{\text{mean}} \, \text{(mm)}\)}} & \multirow{2}{*}{\shortstack{Max. Euclidean Error\(_{\text{mean}}\) \\ (mm)}} \\
        & & & \\
        \hline        
        Hold-out & 46.56 & 24.02 $\pm$ 13.27 & 2.60 $\pm$ 1.40 \\
        
        \shortstack{\ac{LODO}} & 46.56 & 46.56 $\pm$ 0.00 & 5.16 $\pm$ 0.00 \\
        \hline
    \end{tabularx}
    \label{tab:statistics}
\end{table*}

\end{document}